\documentclass{article}
\DeclareUnicodeCharacter{2009}{\,}

 \usepackage[main, final]{neurips_2025}

\usepackage[utf8]{inputenc} 
\usepackage[T1]{fontenc}    
\usepackage{hyperref}       
\usepackage{url}            
\usepackage{booktabs}       
\usepackage{amsfonts}       
\usepackage{nicefrac}       
\usepackage{microtype}      
\usepackage{xcolor}         
\usepackage{multicol}
\usepackage{multirow}
\usepackage{adjustbox}
\usepackage{algorithm}
\usepackage{algpseudocode}
\usepackage{amsmath}
\usepackage{amssymb}
\usepackage{graphicx}
\usepackage{subcaption}     
\usepackage{cleveref}       
\usepackage{placeins}       
\usepackage{todonotes}      
\usepackage{multicol}       
\usepackage{microtype}      
\usepackage{makecell}

\newcommand{\loss}{\mathcal{L}}

\title{Uncertainty-aware Physics-informed Neural Networks for Robust CARS-to-Raman Signal Reconstruction}

\author{
Aishwarya Venkataramanan 
\And Sai Karthikeya Vemuri \And Adithya Ashok Chalain Valapil \And Joachim Denzler \\ Computer Vision Group, 
  Friedrich Schiller University Jena, Germany\\
}

\begin{document}

\maketitle

\begin{abstract}
Coherent anti-Stokes Raman scattering (CARS) spectroscopy is a powerful and rapid technique widely used in medicine, material science, and chemical analyses. 
However, its effectiveness is hindered by the presence of a non-resonant background that interferes with and distorts the true Raman signal. 
Deep learning methods have been employed to reconstruct the true Raman spectrum from measured CARS data using labeled datasets. A more recent development integrates the domain knowledge of Kramers-Kronig relationships and smoothness constraints in the form of physics-informed loss functions. However, these deterministic models lack the ability to quantify uncertainty, an essential feature for reliable deployment in high-stakes scientific and biomedical applications. In this work, we evaluate and compare various uncertainty quantification (UQ) techniques within the context of CARS-to-Raman signal reconstruction. 
Furthermore, we demonstrate that incorporating physics-informed constraints into these models improves their calibration, offering a promising path toward more trustworthy CARS data analysis.
\end{abstract}

\section{Introduction}

Coherent anti-Stokes Raman scattering (CARS) spectroscopy has emerged as an important technique in chemistry, physics, and biomedical imaging due to its high-speed, label-free detection of molecular vibrations~\cite{cars_theory,cars_theory2}. 
Its ability to provide rapid chemical contrast makes it particularly attractive for live-cell imaging and real-time diagnostics. 
However, a fundamental limitation of CARS is the presence of a strong non-resonant background (NRB), which interferes with the resonant Raman response, distorting both the spectral shape and intensity. This reduces the spectral interpretability and imaging contrast~\cite{nrb_theory}.

To address this challenge, recent works use deep learning to learn the mappings from the NRB-contaminated CARS spectra to the underlying true Raman spectra. 
One promising direction is the use of physics-informed neural networks (PINNs)~\cite{Raissi2017,Raissi2019,vemuri2024, vemuri2025rampinnrecoveringramanspectra}, which embed physical constraints directly into the learning process.
Recently, a model named RamPINN was developed by integrating known physical relationships between signal components. 
It is shown to reconstruct Raman spectra that are both data-consistent and physically plausible, improving the separation of resonant and non-resonant components even under challenging measurement conditions~\cite{vemuri2025rampinnrecoveringramanspectra}.
Nevertheless, RamPINN is deterministic in nature and does not account for model uncertainty. This is an increasingly important consideration when deploying machine learning models in domains where decisions have significant consequences, such as in biomedical diagnostics or materials characterization~\cite{gal2016dropout, venkataramanan2023gaussian}.

In this study, we explore the integration of uncertainty quantification (UQ) methods with PINNs for CARS-to-Raman signal reconstruction.
To this end, we systematically evaluate the prominent UQ strategies: Monte Carlo Dropout~\cite{gal2016dropout}, Bayesian Neural Networks (BNNs)~\cite{blundell2015weight, sharma2023bayesian}, Deep Ensembles~\cite{lakshminarayanan2017simple}, and Neural Processes (NPs)~\cite{venkataramanan2025distance}. The results indicate that incorporating physical knowledge consistently improves reconstruction accuracy across the UQ methods and yields better confidence calibration.


\section{Physics-Informed Learning}

\subsection{General Setup}
Physics-informed learning integrates neural networks with governing physical laws to embed domain knowledge directly into the training process~\cite{Raissi2019, Karniadakis2021}.
The goal is to obtain models that not only fit observed data but also satisfy underlying physical constraints.

Let $x$ be the input, $\hat{y}$ the prediction within the domain $\Omega$, and $f_\theta$ a neural network parameterized by $\theta$. A known operator $\mathcal{D}$ encodes the physical constraint:
\begin{equation}
\mathcal{D}(\hat{y}, x) = 0~, \quad x \in \Omega~,
\end{equation}
where $\mathcal{D}$ is typically a differential or integral operator~\cite{Raissi2019}. To enforce this constraint during training, a physics loss term is defined to penalize violations:
\begin{equation}
\loss_{\text{phy}} = \frac{1}{|\Omega|} \int_\Omega \left| \mathcal{D}(f_\theta(x), u(x), x) \right|^2 dx~.
\end{equation}
where $u(x)$ represents any additional known functions or auxiliary variables required by the constraint.
The total training objective combines the physics loss with a standard data-fitting term (e.g., mean squared error):
\begin{equation}
\loss_{\text{total}} = \lambda_{\text{data}} \cdot \loss_{\text{data}} + \lambda_{\text{phy}} \cdot \loss_{\text{phy}}.
\end{equation}
Here, $\lambda_{\text{data}}$ and $\lambda_{\text{phy}}$ control the relative weights of the data and physics components~\cite{vemuri,vemuri2024,mcclenny2022selfadaptive}.

\subsection{Kramers--Kronig and Smoothness Constraints}

Following RamPINN\cite{vemuri2025rampinnrecoveringramanspectra}, let $x$ denote the measured CARS spectrum, and $\hat{y} = (\hat{y}_{\text{raman}}, \hat{y}_{\text{NRB}})$ the predicted Raman and NRB components from $f_\theta(x)$.  
These components are related through the Kramers--Kronig (KK) relations~\cite{kk_book,kramers1928Diffusion,kronig1926dispersion,guenther2004Encyclopedia, vemuri2025rampinnrecoveringramanspectra}.  

\paragraph{Kramers--Kronig Regularization.}  
From the causality principle, the real and imaginary parts of the CARS response are related through the Hilbert transform. Accordingly, the Raman component should correspond to the imaginary part of the Hilbert transform of the CARS signal after removing the non-resonant background (NRB)~\cite{vemuri2025rampinnrecoveringramanspectra}:
\begin{equation}
\mathcal{L}_{\text{KK}} = \left| \hat{y}_{\text{raman}} - \Im\left[\mathcal{H}\left(x - \hat{y}_{\text{NRB}}\right)\right] \right|^2 ,
\end{equation}
where $\mathcal{H}(\cdot)$ denotes the differentiable Hilbert transform.  

\paragraph{NRB Smoothness Regularization.}  
The NRB component is expected to be broad and smooth~\cite{nrb_smooth,junjuri2021bidirectional,wang2021vector}.  
To enforce this, we penalize sharp variations:
\begin{equation}
\mathcal{L}_{\text{smooth}} = \left| \nabla \hat{y}_{\text{NRB}} \right|^2 .
\end{equation}

\paragraph{Final Loss.}  
The final objective combines the data fidelity term with both regularization terms:
\begin{equation}
\mathcal{L}_{\text{total}} =
\lambda_{\text{data}} \, \mathcal{L}_{\text{data}} +
\lambda_{\text{KK}} \, \mathcal{L}_{\text{KK}} +
\lambda_{\text{smooth}} \, \mathcal{L}_{\text{smooth}} ,
\label{eq:loss_total}
\end{equation}
where $\lambda_{\text{data}}$, $\lambda_{\text{KK}}$, and $\lambda_{\text{smooth}}$ control the relative contributions of each component.

\section{Uncertainty Quantification}
Unlike deterministic approaches, UQ enables models to express predictive confidence, providing a principled measure of reliability. In this work, we evaluate six Bayesian and ensemble-based UQ methods: Gaussian Processes (GP)~\cite{williams2006gaussian},  Bayesian Neural Networks (Full BNN), Partial BNNs~\cite{sharma2023bayesian}, Monte Carlo Dropout (MC-Dropout)\cite{gal2016dropout}, Deep Ensembles\cite{lakshminarayanan2017simple}, and Distance-informed Neural Processes (DNP)~\cite{venkataramanan2025distance}.
Among these, GP serves as a non-neural baseline and is a probabilistic model with closed-form uncertainty estimates. The remaining methods are neural network-based. BNNs place Gaussian distributions over the network weights and use variational inference to approximate the full posterior. However, these methods are typically not scalable to large or deep networks due to the high computational cost of posterior approximation. Partial BNNs reduce this complexity by applying the Bayesian treatment to only the final layer of the model, offering a trade-off between scalability and uncertainty quality. MC-Dropout approximates Bayesian inference by performing multiple stochastic forward passes at test time, utilizing dropout, which makes it computationally efficient and easy to implement. Deep Ensembles train multiple independently initialized models and combine their predictions to estimate uncertainty. Finally, DNP extends Neural Processes by incorporating input-dependent uncertainty through distance-aware mechanisms, enabling flexible and structured uncertainty modeling.

To investigate the impact of physics-informed inductive biases on uncertainty estimates, we train each neural network-based model both with and without the physics-based loss described in \autoref{eq:loss_total}. In this setup, the data loss term $\mathcal{L}_{\text{data}}$ corresponds to the standard objective traditionally used to train the respective UQ model. The physics-informed training extends this baseline by incorporating additional constraints: Kramers-Kronig consistency loss $\mathcal{L}_{\text{KK}}$ and the NRB smoothness loss $\mathcal{L}_{\text{smooth}}$ to embed domain knowledge into the learning process.

\section{Experiments and Results}\label{sec:exp}
\subsection{Experimental Setup}
\paragraph{Synthetic spectra generation.} We generate 2000 CARS-Raman spectral pairs for our experiments. The spectral synthesis~\cite{specnet} and specific parameters for generation are as in~\cite{valapil2025dadmd}. The synthetic Raman spectra with $N$ peaks are modeled using a Lorentzian function as shown in~\autoref{eq:res}. Here, $A$ is the amplitude, $\Omega$ is the normalized resonance frequency, $\gamma$ is the linewidth and \( \omega \) is the normalized Raman shift:
\begin{equation}
\label{eq:res}
    \chi_r^{(3)}(\omega) = \sum_{n=1}^{N} \frac{A_n}{\Omega_n - \omega - i\gamma_n}.
\end{equation}

A polynomial or sigmoid function is used to replicate the NRB as shown in~\autoref{eq:nrb}. Here, $b_1, b_2$ control the steepness while $c_1, c_2$ determine the position of inflection points in the sigmoid function. Variables $a, b, c, d, e$ are the coefficients of polynomial function:  
\begin{equation}
    \label{eq:nrb}
    \chi_{nrb}^{(3)}(\omega) =  \left\{\begin{matrix}
 \frac{1}{1 + e^{-b_1(\omega - c_1)}} \cdot \frac{1}{1 + e^{b_2(\omega - c_2)}}\, & \text{(Sigmoid)}  \\ 
 a\omega^4 + b\omega^3 + c\omega^2 + d\omega + e\,& \text{(Polynomial)}  \\
\end{matrix}\right..
\end{equation}
This background along with some random noise $\epsilon$ is added over the Raman spectra to generate CARS spectra:
\begin{equation}
    \label{eq:cars1}
    I_{\text{CARS}}(\omega) = \left| \chi_{r}^{(3)}(\omega) + \chi^{(3)}_{\text{nrb}}(\omega) \right|^2\, + \epsilon.
\end{equation}

\paragraph{Real spectra.} We use six homogeneous sample spectra in our experiments. They are measured on broadband CARS and Spontaneous Raman setups, and are publicly available from the work~\cite{real_sample,realdata_zenodo}. 

\paragraph{Training and Evaluation.}
For synthetic experiments, we randomly split the generated dataset into 80\% for training and 20\% for evaluation. For real-world experiments, we adopt a zero-shot evaluation protocol, where the models trained on the synthetic dataset and directly evaluated on real CARS-Raman measurements without any fine-tuning. The neural network backbone follows a 1D ResNet-style architecture consisting of four residual blocks~\cite{he2016deep}. These shared features are then passed to two separate 
$1 \times 1$ convolutional heads for predicting the resonant Raman component and the NRB. We train all models in PyTorch~\cite{paszke2019pytorch} using the Adam optimizer~\cite{kingma2014adam} with a learning rate of $10^{-3}$. The weight terms for the loss function are $\lambda_\text{data} = 10$, $\lambda_\text{KK}=1$ and $\lambda_\text{smooth}=10$. The MC-Dropout results were computed using 50 stochastic forward passes, while the deep ensemble results were obtained from an ensemble of 5 independently trained models.

For quantitative evaluation, we use the log-likelihood (LL) and expected calibration error (ECE)~\cite{kuleshov2018accurate}. LL measures how well the predicted probability distribution explains the observed data, with higher values indicating more reliable uncertainty modeling. ECE quantifies the deviation between predicted confidence intervals and actual coverage. A lower value of ECE indicates a better performance.

\subsection{Synthetic data experiments}
\autoref{tab:synthetic} presents the quantitative comparison of various UQ methods on the synthetic dataset, evaluated both with and without physics-based constraint. Among all methods, the Full BNN achieves the best performance in both settings. It obtains the highest LL and lowest ECE, indicating accurate and well-calibrated predictive uncertainties. The DNP model performs competitively, closely matching the Full BNN. 
The incorporation of physics-based loss consistently improves performance across most neural-based models. 
These results suggest that physics-informed training improves both the predictive performance of the models and calibration of uncertainty estimates for neural network-based approaches. The qualitative results are provided in \autoref{fig:synthetic} in \autoref{app:qualitative}.

\begin{table}[]
\caption{Comparison of uncertainty quantification methods on a synthetic dataset, evaluated with and without physics. Results are averaged over 10 runs. }
\begin{adjustbox}{width=0.7\columnwidth, center}
    \centering
    \begin{tabular}{ccccc} \toprule
       \multirow{2}{*}{Method} & \multicolumn{2}{c}{Without Physics} & \multicolumn{2}{c}{With Physics} \\
       \cmidrule(lr){2-3} \cmidrule(lr){4-5}
       &  LL ($\uparrow$) & ECE ($\downarrow$)  & LL ($\uparrow$) & ECE ($\downarrow$)   \\ \midrule
       GP & 1.115$\pm$0.017 & 0.202$\pm$0.004 &  -  & -\\
       MC-Dropout  & -1.164$\pm$0.013 & 0.213$\pm$0.015  & -2.089$\pm$0.053 & 0.278$\pm$0.012 \\
       Deep Ensemble & -1.102$\pm$0.015 & 0.215$\pm$0.005  & 0.119$\pm$0.023 & 0.145$\pm$0.006 \\
       Full BNN & \textbf{1.189$\pm$0.046} & \textbf{0.076$\pm$0.003}  & \textbf{1.274$\pm$0.076} & \textbf{0.049$\pm$0.004}  \\
       Partial BNN & 0.731$\pm$0.012 & 0.280$\pm$0.003  & 0.762$\pm$0.004 & 0.234$\pm$0.007 \\
       DNP & \underline{1.175$\pm$0.021} & \underline{0.114$\pm$0.003} & \underline{1.188$\pm$0.017} & \underline{0.104$\pm$0.005}  \\ \bottomrule
    \end{tabular}
    \label{tab:synthetic}
\end{adjustbox}
\end{table}

\subsection{Real-world experiments}

\autoref{tab:real} summarizes the performance of the UQ baselines on the real CARS-Raman dataset under zero-shot setting. DNP achieves the best overall performance, yielding the highest LL and lowest ECE both with and without physics-based regularization. This strong performance can be attributed to DNP’s meta-learning architecture, which is explicitly designed to enable fast adaptation and generalization across tasks. Full BNN also performs competitively, showing strong LL scores and improved calibration when physics-informed loss terms are included. Similar to the observation from the experiments on synthetic dataset, incorporating physics improves both metrics for most models. This suggests that domain knowledge can help reduce overconfidence and improve generalization even in zero-shot scenarios. Qualitative results are provided in \autoref{fig:real} in \autoref{app:qualitative}.

\begin{table}[]
    \caption{Comparison of uncertainty quantification methods on a real dataset, evaluated with and without physics. Results are averaged over 10 runs.}
    \begin{adjustbox}{width=0.7\columnwidth, center}
    \centering
    \begin{tabular}{ccccc} \toprule
       \multirow{2}{*}{Method} & \multicolumn{2}{c}{Without Physics} & \multicolumn{2}{c}{With Physics} \\
       \cmidrule(lr){2-3} \cmidrule(lr){4-5}
       &  LL ($\uparrow$) & ECE ($\downarrow$)  & LL ($\uparrow$) & ECE ($\downarrow$)   \\ \midrule
       GP & \underline{0.685$\pm$0.164} & \textbf{0.087$\pm$0.049}  & - & -  \\
       MC-Dropout  & -3.089$\pm$0.530 & 0.278$\pm$0.012  & -3.530
       $\pm$0.598 & 0.242$\pm$0.009 \\
       Deep Ensemble  & -2.567$\pm$0.436 & 0.255$\pm$0.021 &  -2.016$\pm$0.418 & 0.208$\pm$0.012 \\
       Full BNN & \underline{0.685$\pm$0.231} & 0.172$\pm$0.041  & \underline{0.716$\pm$0.263} & \underline{0.160$\pm$0.076} \\
       Partial BNN & 0.497$\pm$0.172 & 0.227$\pm$0.021  & 0.565$\pm$0.188 & 0.215$\pm$0.027 \\
       DNP & \textbf{0.914$\pm$0.185} & \underline{0.168$\pm$0.062} & \textbf{1.014$\pm$0.144} & \textbf{0.131$\pm$0.064} \\ \bottomrule
    \end{tabular}
    \label{tab:real}
\end{adjustbox}
\end{table}

\section{Conclusion}\label{sec:conclusion}

In this study, we conducted a systematic evaluation of UQ methods for CARS-to-Raman signal reconstruction. Experiments on synthetic and real datasets show that embedding domain knowledge via physics consistently improves reconstruction accuracy and uncertainty calibration. Among the evaluated UQ methods, Full BNNs and Distance-informed Neural Processes achieved the best overall performance, with physics-informed training yielding higher log-likelihood scores and lower calibration errors. More broadly, these results demonstrate that integrating physical priors not only improves signal fidelity but also enhances the reliability and interpretability of uncertainty estimates. Future work will explore scaling these approaches to broader spectral ranges, multi-modal imaging, and active learning frameworks that leverage uncertainty estimates for adaptive data acquisition.

\bibliographystyle{unsrt}
\bibliography{references}

\newpage

\appendix

\section{Qualitative Evaluation}\label{app:qualitative}

\begin{figure}[h]
    \centering
    \begin{tabular}{cc} 
        & 
        \begin{minipage}[c]{0.9\linewidth} 
            \rotatebox{90}{\hspace{3em}\small Input}
            \includegraphics[width=\linewidth]{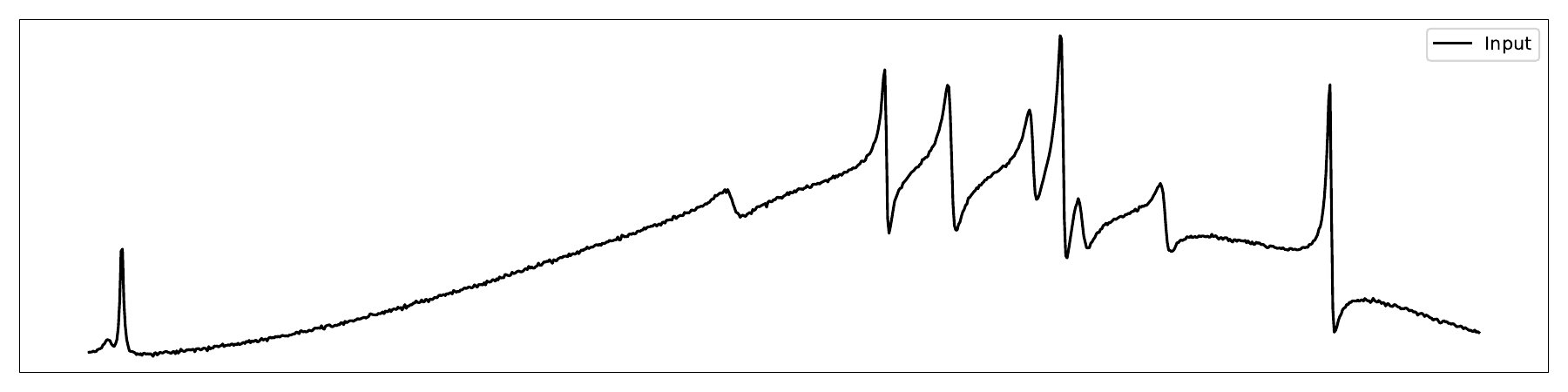}
        \end{minipage} \\[0.3cm] 

        \multirow{2}{*}{\rotatebox{90}{\hspace{4em}\large Without Physics}} & 
        \begin{minipage}[c]{0.9\linewidth}
            \rotatebox{90}{\hspace{3em}\small Full BNN}
            \includegraphics[width=\linewidth]{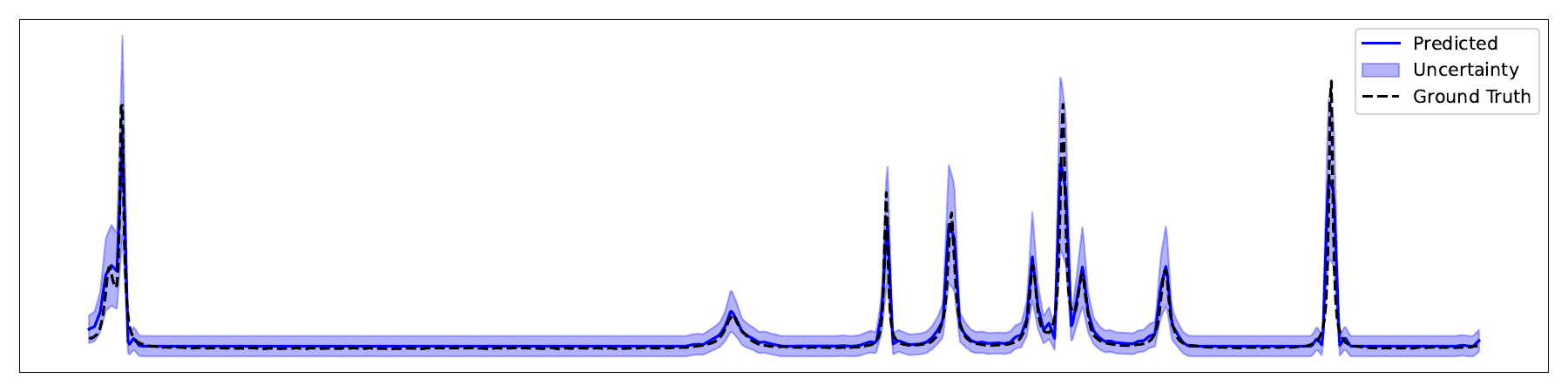}
        \end{minipage} \\[0.3cm]

        & 
        \begin{minipage}[c]{0.9\linewidth}
            \rotatebox{90}{\hspace{1.5cm}\small DNP}
            \includegraphics[width=\linewidth]{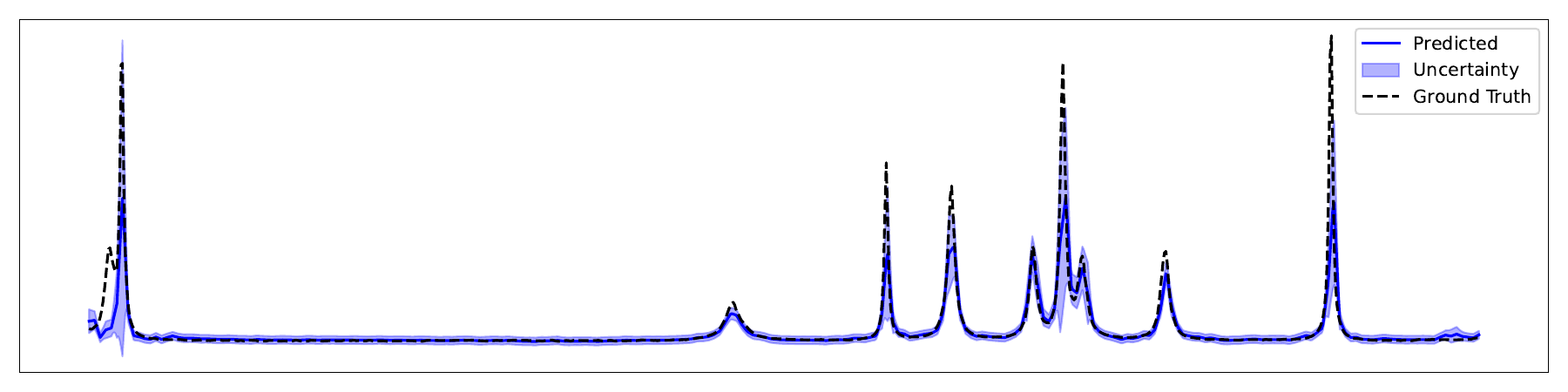}
        \end{minipage} \\[0.3cm]

        \multirow{2}{*}{\rotatebox{90}{\hspace{5em}\large With Physics}} & 
        \begin{minipage}[c]{0.9\linewidth}
            \rotatebox{90}{\hspace{3em}\small Full BNN}
            \includegraphics[width=\linewidth]{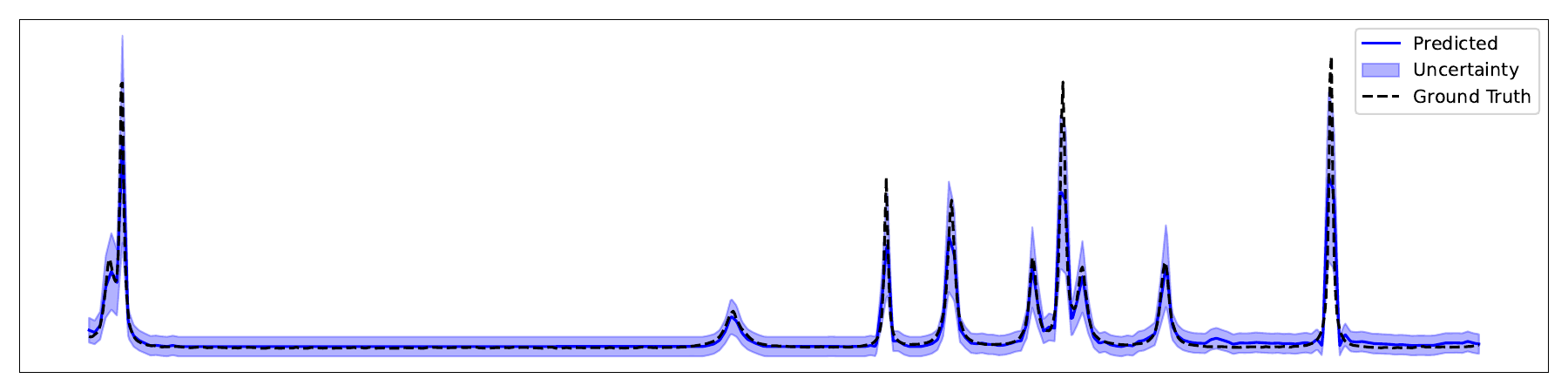}
        \end{minipage} \\[0.3cm]
        
        & 
        \begin{minipage}[c]{0.9\linewidth}
            \rotatebox{90}{\hspace{1.5cm}\small DNP}
            \includegraphics[width=\linewidth]{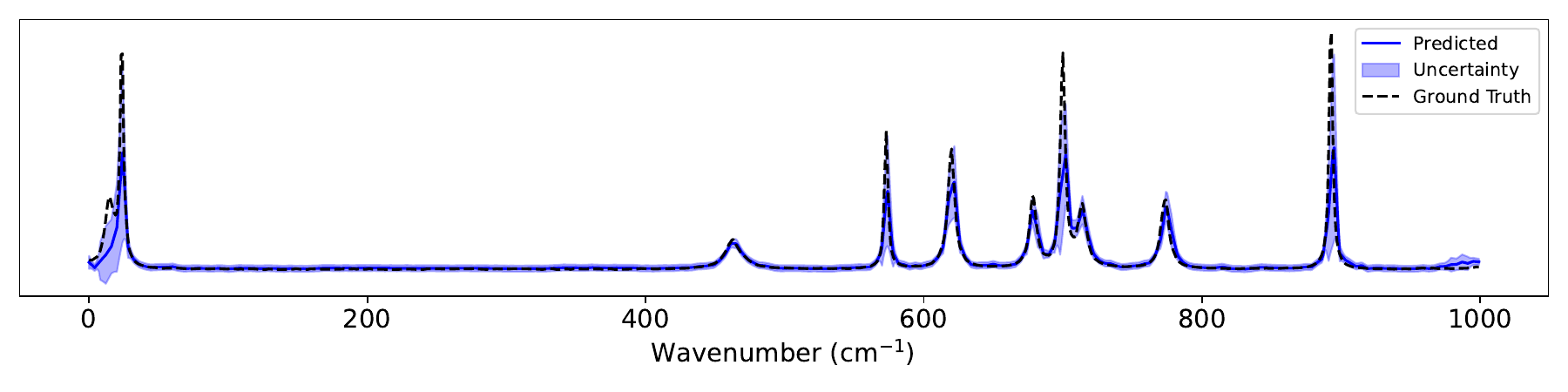}
        \end{minipage}
    \end{tabular}
    \caption{\textbf{Synthetic dataset results.} Reconstructions from the Full BNN and DNP models without (top) and with (bottom) the physics constraint.}
    \label{fig:synthetic}
\end{figure}

\begin{figure}[h]
    \centering
    \begin{tabular}{cc} 
        & 
        \begin{minipage}[c]{0.9\linewidth} 
            \rotatebox{90}{\hspace{3em}\small Input}
            \includegraphics[width=\linewidth]{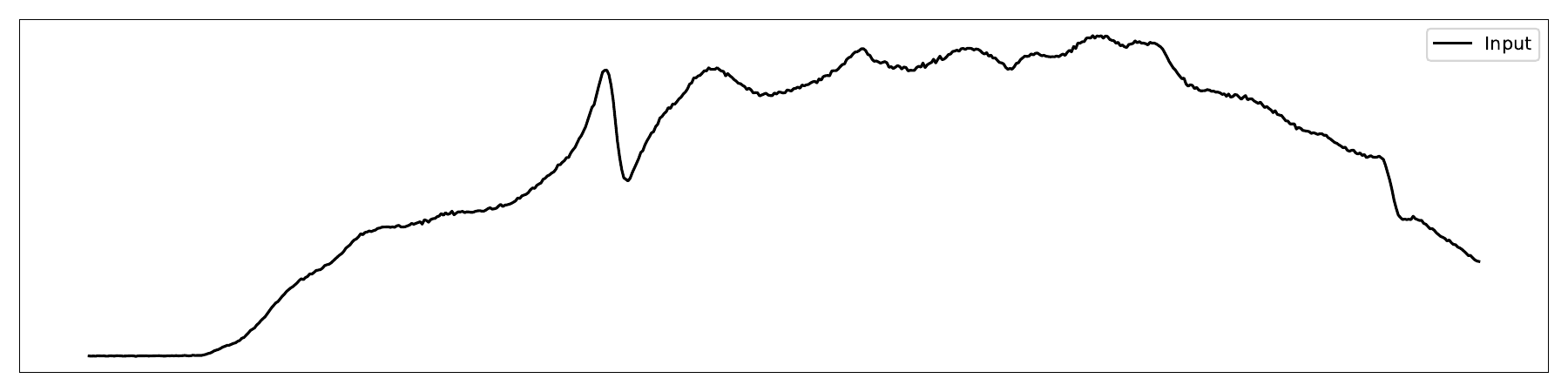}
        \end{minipage} \\[0.3cm] 

        \multirow{2}{*}{\rotatebox{90}{\hspace{4em}\large Without Physics}} & 
        \begin{minipage}[c]{0.9\linewidth}
            \rotatebox{90}{\hspace{3em}\small Full BNN}
            \includegraphics[width=\linewidth]{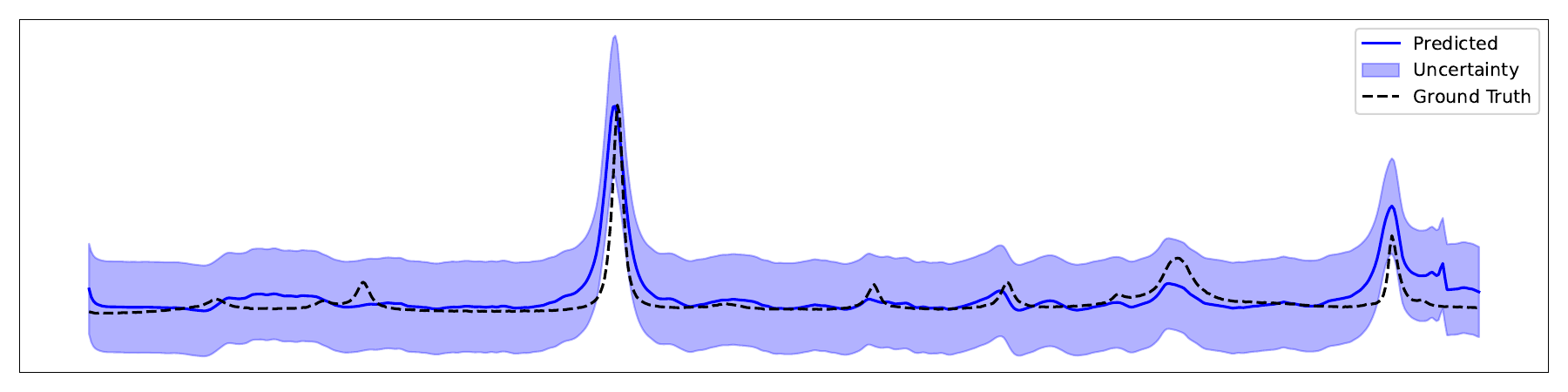}
        \end{minipage} \\[0.3cm]

        & 
        \begin{minipage}[c]{0.9\linewidth}
            \rotatebox{90}{\hspace{1.5cm}\small DNP}
            \includegraphics[width=\linewidth]{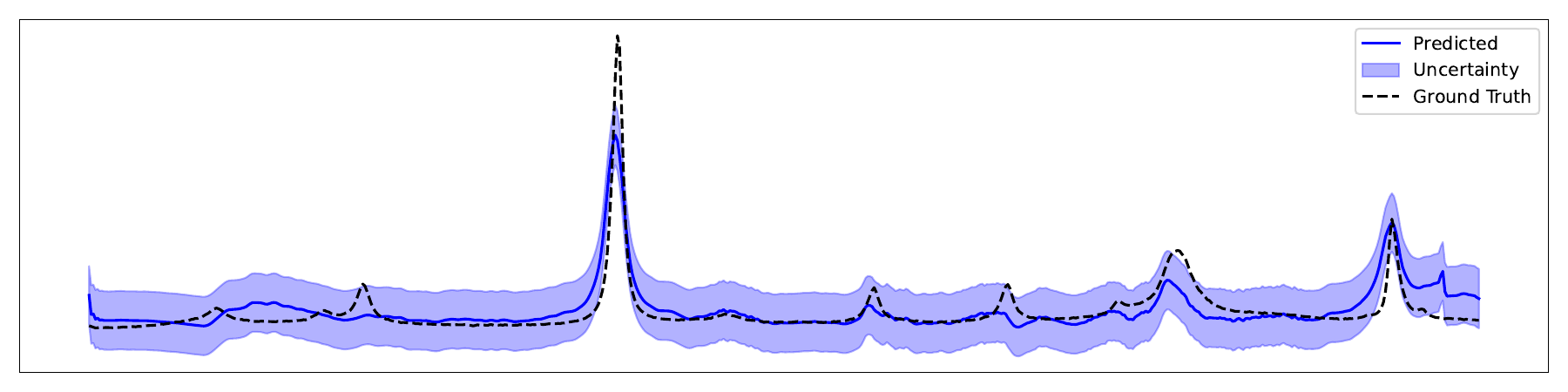}
        \end{minipage} \\[0.3cm]

        \multirow{2}{*}{\rotatebox{90}{\hspace{5em}\large With Physics}} & 
        \begin{minipage}[c]{0.9\linewidth}
            \rotatebox{90}{\hspace{3em}\small Full BNN}
            \includegraphics[width=\linewidth]{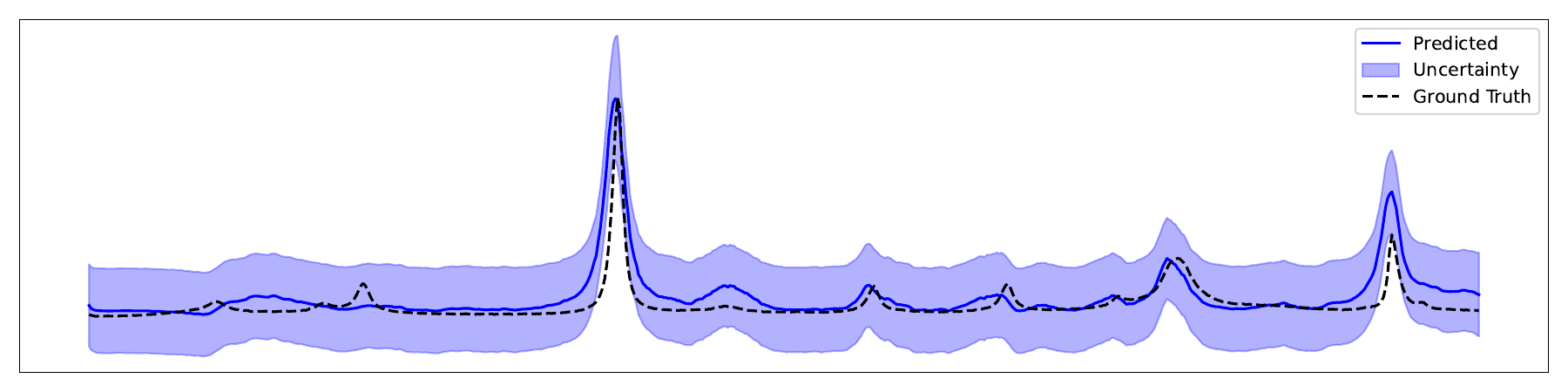}
        \end{minipage} \\[0.3cm]
        
        & 
        \begin{minipage}[c]{0.9\linewidth}
            \rotatebox{90}{\hspace{1.5cm}\small DNP}
            \includegraphics[width=\linewidth]{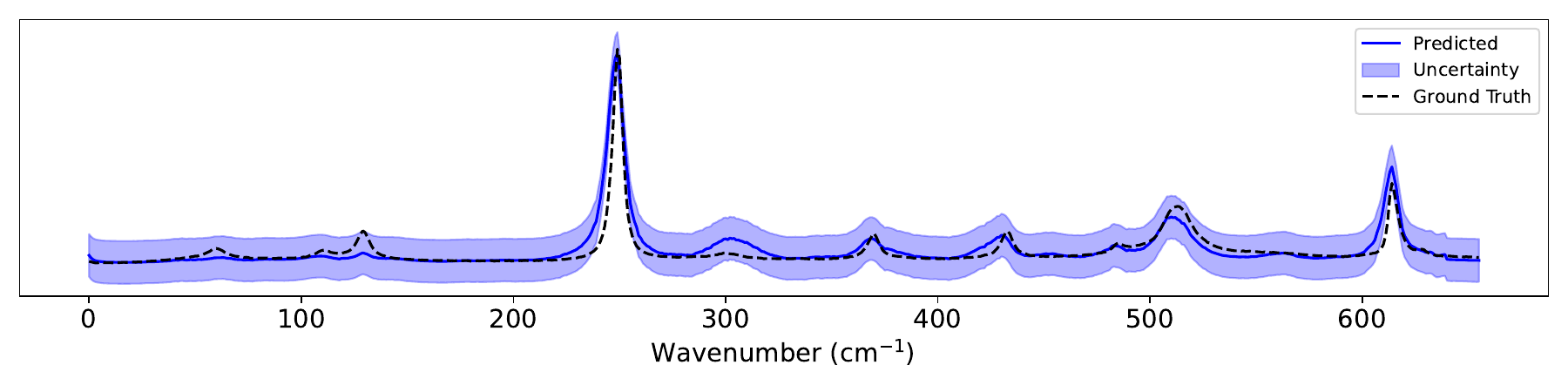}
        \end{minipage}
    \end{tabular}
    \caption{\textbf{Real dataset results.} Zero-shot reconstructions from the Full BNN and DNP models without (top) and with (bottom) the physics constraint.}
    \label{fig:real}
\end{figure}

We qualitatively assess the impact of physics-informed constraints on the Full BNN and DNP models. \autoref{fig:synthetic} illustrates their performance on a representative sample from the synthetic dataset, while \autoref{fig:real} presents results on the real dataset.

In the synthetic case (\autoref{fig:synthetic}), all models closely reproduce the ground-truth Raman spectra and effectively suppress the non-resonant background. Incorporating the physics constraint further reduces reconstruction errors. The models, however, exhibit distinct uncertainty behaviors: the Full BNN shows broadly distributed uncertainty across the spectrum, whereas the DNP yields smoother, more localized estimates concentrated in low-signal-to-noise regions.

For the real dataset (\autoref{fig:real}), both methods generalize well in a zero-shot setting, with the DNP maintaining more consistent uncertainty calibration. Consistent with the synthetic results, applying the physics constraint again leads to lower reconstruction errors and improved reliability.


\end{document}